%% file: main.tex
\documentclass[sigconf]{acmart}

\AtBeginDocument{%
  \providecommand\BibTeX{{%
    \normalfont B\kern-0.5em{\scshape i\kern-0.25em b}\kern-0.8em\TeX}}}

\copyrightyear{2022}
\acmYear{2022}
\setcopyright{acmlicensed}\acmConference[SIGIR '22]{Proceedings of the 45th
International ACM SIGIR Conference on Research and Development in
Information Retrieval}{July 11--15, 2022}{Madrid, Spain}
\acmBooktitle{Proceedings of the 45th International ACM SIGIR Conference on
Research and Development in Information Retrieval (SIGIR '22), July 11--15,
2022, Madrid, Spain}
\acmPrice{15.00}
\acmDOI{10.1145/3477495.3531757}
\acmISBN{978-1-4503-8732-3/22/07}

\usepackage{amsmath}
\usepackage{booktabs}
\usepackage{enumitem}
\usepackage{multirow}

\newcommand{\model}{MorsE}

\begin{document}

\title{
Meta-Knowledge Transfer for Inductive Knowledge Graph Embedding 
}


\author{Mingyang Chen}
\authornote{Equal Contribution.}
\email{mingyangchen@zju.edu.cn}
\affiliation{%
  \institution{College of Computer Science and Technology, Zhejiang University}
  \country{}
}

\author{Wen Zhang}
\email{wenzhang2015@zju.edu.cn}
\authornotemark[1]
\affiliation{%
  \institution{School of Software Technology, Zhejiang University}
  \country{}
}

\author{Yushan Zhu}
\author{Hongting Zhou}
\email{yushanzhu@zju.edu.cn}
\email{seven777@zju.edu.cn}
\affiliation{%
  \institution{College of Computer Science and Technology, Zhejiang University}
  \country{}
}


\author{Zonggang Yuan}
\email{yuanzonggang@huawei.com}
\affiliation{%
  \institution{Huawei Technologies Co., Ltd.}
  \country{}
}

\author{Changliang Xu}
\email{xu@shuwen.com}
\affiliation{%
  \institution{State Key Laboratory of Media Convergence Production Technology and Systems}
  \city{Beijing}
  \country{China}
  \country{}
}

\author{Huajun Chen}
\authornote{Corresponding author.}
\email{huajunsir@zju.edu.cn}
\affiliation{%
    \institution{College of Computer Science and Technology, Zhejiang University}
    \institution{ZJU-Hangzhou Global Scientific and Technological Innovation Center}
    \institution{Alibaba-Zhejiang University Joint Institute of Frontier Technologies}
  \country{}
}

\fancyhead{}

\begin{abstract}
Knowledge graphs (KGs) consisting of a large number of triples have become widespread recently, and many knowledge graph embedding (KGE) methods are proposed to embed entities and relations of a KG into continuous vector spaces.
Such embedding methods simplify the operations of conducting various in-KG tasks (e.g., link prediction) and out-of-KG tasks (e.g., question answering). They can be viewed as general solutions for representing KGs.
However, existing KGE methods are not applicable to inductive settings, where a model trained on source KGs will be tested on target KGs with entities unseen during model training.
Existing works focusing on KGs in inductive settings can only solve the inductive relation prediction task. They can not handle other out-of-KG tasks as general as KGE methods since they don't produce embeddings for entities.
In this paper, to achieve inductive knowledge graph embedding, we propose a model \textbf{\model}, which does not learn embeddings for entities but learns transferable \textit{meta-knowledge} that can be used to produce entity embeddings.
Such meta-knowledge is modeled by entity-independent modules and learned by meta-learning.
Experimental results show that our model significantly outperforms corresponding baselines for in-KG and out-of-KG tasks in inductive settings\footnote{Source code is available at \url{https://github.com/zjukg/MorsE}.}.
\end{abstract}




\maketitle

\section{Introduction}

Knowledge graphs (KGs) consist of a large number of facts in the form of triples like \textit{(head entity, relation, tail entity)}, and have benefited a lot of downstream tasks recently.
Nowadays many large-scale knowledge graphs, including Freebase \cite{freebase}, NELL \cite{nell}, and Wikidata \cite{wikidata}, have been proposed and have supported many in-KG applications, e.g., link prediction \cite{TransE} and triple classification \cite{TransH}, as well as out-of-KG applications, e.g., question answering \cite{QA-GNN} and recommender systems \cite{recsys, AliCG}.
\begin{figure}[t]
\centering
\includegraphics[scale=0.41]{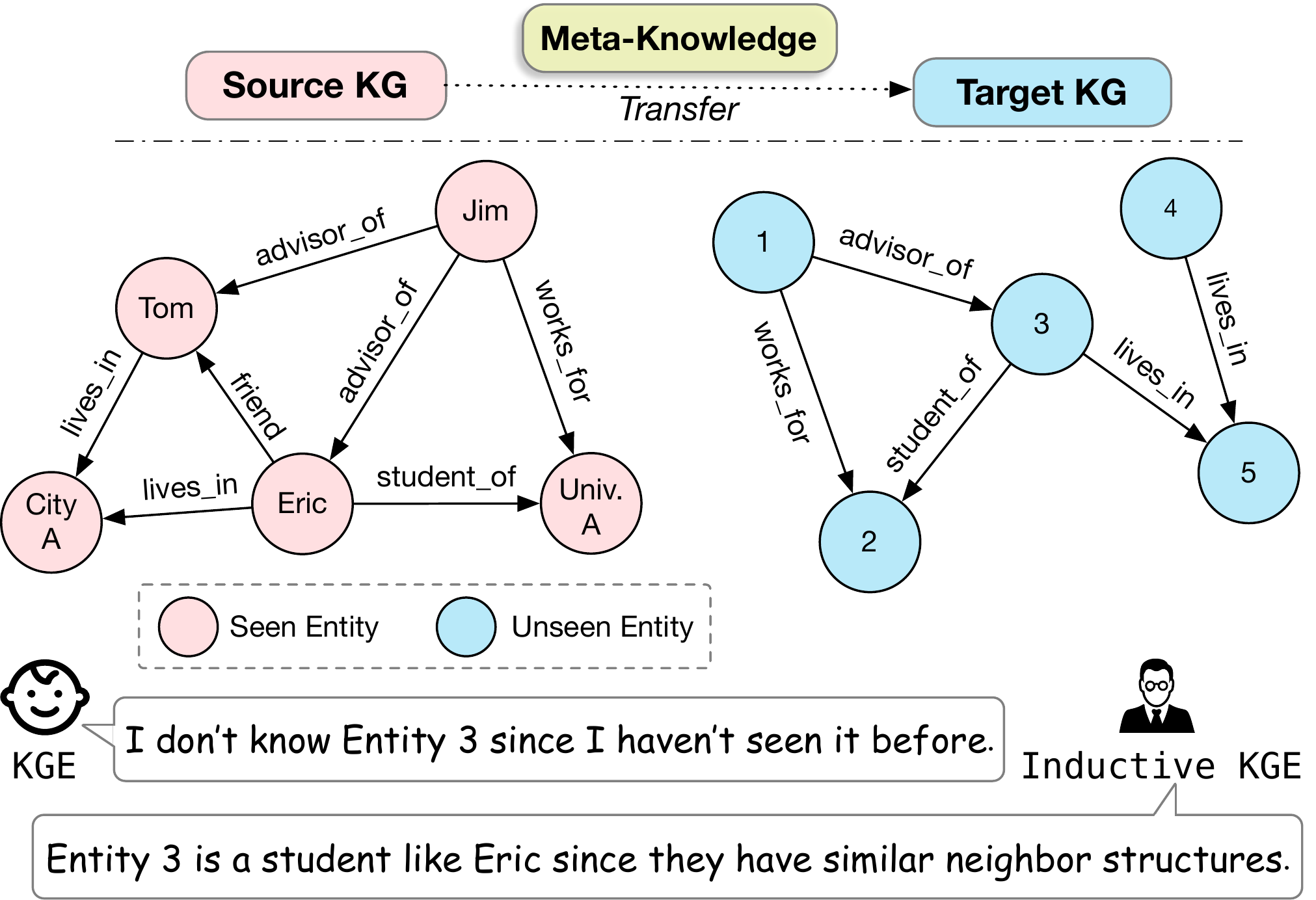}
\caption{An example of KGs in the inductive setting. A conventional KGE method is like a baby who can only recognize seen entities, while the inductive KGE can recognize an unseen entity by transferable structural patterns.}
\label{fig:intro_v2}
\end{figure}

It is hard to use discrete structured triples directly when applying knowledge graphs in various applications since modern deep learning methods can not easily manipulate such representations. 
Hence, many recent works embed entities and relations of a knowledge graph into continuous vector spaces, such as TransE \cite{TransE}, ComplEx \cite{ComplEx}, and RotatE \cite{RotatE}. 
Learned embeddings (i.e., vector representations) for knowledge graphs can be used for in-KG tasks to improve the quality of KGs and out-of-KG tasks to help inject background knowledge such as similarity and relational information between entities.
Thus knowledge graph embedding (KGE) models are general solutions to represent knowledge graphs, and embeddings can preserve the inherent semantics of a KG \cite{kgembedding}.

However, in conventional knowledge graph embedding methods, promising predictions and applications are only ensured for entities seen during training, not unseen entities, since they learn embeddings for a fixed predefined set of entities. 
Thus compared to the \textit{transductive} setting where a model is tested on a subset of the training entity set, the \textit{inductive} setting is more challenging for KGE methods where new entities appear while testing. 
For example, as shown in Figure~\ref{fig:intro_v2}, a KGE model trained on the source KG can not be used on the target KG since the entities in the target KG are not seen during model training on the source KG.

Recently, significant progress has been achieved in inductive settings for KGs, especially for in-KG tasks. 
Specifically, GraIL \cite{GraIL}, and its following works \cite{compile, TACT} are proposed to conduct inductive relation prediction. They solve this problem by learning to predict relations from the subgraph structure around a candidate relation, avoiding learning embedding for any specific entity.
Although these methods proved adequate for the inductive relation prediction task, they can not solve other out-of-KG tasks since they don't produce embeddings for entities.
This drawback makes such methods not as general as KGE methods to solve various tasks related to KGs.
To generally solve various tasks for KGs in inductive settings, we raise a research question: \textit{Can we train a knowledge graph embedding model on a set of entities that can generalize to unseen entities for either in-KG or out-of-KG tasks?} 
We refer to this problem as inductive knowledge graph embedding.

To solve this problem, we resort to the process of human cognition. 
As shown in Figure~\ref{fig:intro_v2}, a conventional KGE method is like a baby who can only cognize the entity they have seen. In contrast, an adult can cognize a new entity by comparing its neighbor's structural pattern with seen entities. 
Such structural patterns, which help humans understand the semantics of a new entity, are universal, entity-independent, and transferable.
In this paper, we collectively refer to the knowledge about such transferable structural patterns as \textit{meta-knowledge}.
It is the capability of modeling and learning such meta-knowledge making adults and models can successfully handle KGs in inductive settings.




Inspired by this, in our paper, we propose a novel model, \underline{M}eta-Kn\underline{o}wledge T\underline{r}an\underline{s}fer for Inductive Knowledge Graph \underline{E}mbedding (\textbf{\model}), which can produce high-quality embeddings for new entities in the inductive setting via modeling and learning entity-independent meta-knowledge.

For meta-knowledge modeling,
we focus on modeling the procedure of producing entity embeddings from entity-independent information.
Specifically, we instantiate meta-knowledge as two modules, an entity initializer and a Graph Neural Network (GNN) modulator.
The entity initializer 
initializes each entity embedding via two entity-independent embeddings, including a relation-domain embedding and a relation-range embedding.
The GNN modulator enhances entity embeddings based on entities' neighbor structure information. 
Overall, the parameters for these two modules are independent of any specific entities and can be adapted to KGs with unseen entities after model training. 

For meta-knowledge learning,
we resort to meta-learning to achieve learning to produce entity embeddings. 
Following the meta-training regime, during model training, we sample a set of tasks on the source KG, where each of them consists of a support (triple) set and a query (triple) set. Moreover, entities in these tasks are treated as unseen to simulate the inductive setting for KGs.
For each task, entity embeddings are produced by the entity initializer and the GNN modulator based on the support set and are evaluated on the query set.
We meta-train \model~over these sampled tasks on source KGs to enforce it on producing reasonable and effective embeddings given the support triples of a task. Thus \model~can generalize to target KGs with unseen entities.

We evaluate \model~by inductive settings for in-KG tasks (i.e., link prediction) and out-of-KG tasks (i.e., question answering). 
The results show that our model outperforms other baselines and can effectively achieve meta-knowledge transfer for inductive knowledge graph embedding. The main contributions of our paper are summarized as follows:
\begin{itemize}
\item We emphasize knowledge graph embedding as a general solution to represent KGs for in-KG and out-of-KG tasks, and extend it to inductive settings.
\item We propose a general inductive knowledge graph embedding framework \model~via meta-knowledge transfer and learn such meta-knowledge by meta-learning.
\item We do extensive experiments on both in-KG and out-of-KG tasks for KGs in inductive settings, demonstrating the effectiveness of our model.
\end{itemize}

\section{Related Work}

\subsection{Knowledge Graph Embedding}

Knowledge graph embedding methods learn embeddings for entities and relations while preserving the inherent semantics of KGs \cite{kgembedding}. Such embeddings can alleviate the drawbacks of representing a KG as structured triples and can be easily deployed to many in-KG tasks, including link prediction \cite{TransE, WRAN} and triple classification, and out-of-KG tasks, including question answering \cite{qa} and recommender systems \cite{recsys}. Moreover, KGE methods are also investigated in various scenarios, like knowledge distillation \cite{DualDE} and federated learning \cite{FedE}.

A majority of works on KGE focus on designing expressive score functions to model the triples in a KG \cite{TransD, CrossE, TransH, TransR, IterE}, based on some specific assumption of relational patterns.
Translational distance models measure the plausibility of a triple as the distance between the tail entity embedding and the head entity embedding after a translation carried out by the relation.
TransE \cite{TransE} is a representative translational distance model that represents entities and relations as vectors in the same space and assumes that the relation is a translation vector between entities. Recently, RotatE \cite{RotatE} treats each relation as a rotation from the head entity to the tail entity in the complex vector space. 
Furthermore, semantic matching models design score functions by matching latent semantics of entities and relations. DistMult \cite{DistMult} scores a triple by capturing pairwise interactions between the components of entities along the same dimension. ComplEx \cite{ComplEx} follows DistMult but embeds components of KGs as complex-valued embeddings to model asymmetric relations.

Recently, some work such as R-GCN \cite{RGCN} and CompGCN \cite{CompGCN} adapt GNNs, which are usually used for simple undirected graphs, 
to multi-relational knowledge graphs,
for learning more expressive knowledge graph embeddings.
GNNs can be viewed as encoder models for entities, which encode the features for entities to their embeddings based on entities' multi-hop neighbor structures. GNNs are biased toward graph symmetries and proved effective to capture universal and general structure patterns \cite{GCC}, which could help encode better knowledge graph embeddings equipped with conventional KGE methods as decoders.

Although many KGE methods have been proposed and proved effective in handling various relational
patterns using sophisticated score functions or GNNs, they are designed for KG-related tasks in transductive settings. Namely, they can not handle tasks related to entities unseen during training.

\subsection{Knowledge Graph in Inductive Settings}

The inductive setting for knowledge graphs is a realistic scenario since knowledge graphs are evolving, and the construction of new KGs with new entities is ongoing every day. Various methods are proposed to solve different problems in inductive settings for KGs. Rule-learning-based methods learn probabilistic logical rules, and such rules can be used for inductive reasoning on KG with unseen entities. Specifically, AMIE \cite{AMIE} and RuleN \cite{rulen} learn logical rules explicitly via discrete search. Neural LP \cite{neural-lp} and DRUM \cite{drum} extract rules in an end-to-end differentiable manner. Since rules are independent of specific entities, rule-learning-based methods can handle link prediction with unseen entities. 

Furthermore, some methods resort to additional information for entities like textual description as representations for unseen entities. KG-BERT \cite{KG-BERT} and StAR \cite{star} use pre-trained language models \cite{roberta} to encode textual descriptions for each entity. These methods are inherently inductive since they can produce embeddings for any entities as long as they have corresponding descriptions. However, we don't consider any features or textual information for entities. In our work, the entity embeddings are totally based on the KG structure, which is a more general scenario and can be adapted to various applications.

Some other works focus on handling out-of-knowledge-base entities connected to the trained KG. 
\cite{LAN} and \cite{OOKB} train neighborhood aggregators to embed new entities via their existing neighbors. GEN \cite{GEN} and HRFN \cite{hyper-relation-ookb} learn entities embeddings for both seen-to-unseen and unseen-to-unseen link prediction based on meta-learning.
However, they can not solve our paper's scenario that focuses on handling unseen entities in an entire new KG in the inductive setting.

The most relevant works for our paper are those focused on inductive knowledge graph completion. Such works can generalize to unseen entities and conduct relation prediction on KGs with entirely new entities.
Specifically, GraIL \cite{GraIL}, CoMPILE \cite{compile}, and TACT \cite{TACT} learn the ability of relation prediction by subgraph extraction and GNNs independent of any specific entities. 
Such ability can generalize to unseen entities in completely new KGs. 
Recently, INDIGO \cite{INDIGO} has been proposed to accomplish inductive knowledge graph completion based on a GNN using pair-wise encoding.
Although these methods are adequate for inductive knowledge graph completion tasks, they can not handle other out-of-KG tasks in the inductive setting as general as our model.

\subsection{Meta-Learning}

Meta-learning methods devote to training models that can adapt to new tasks quickly with the concept of ``learning to learn''. Generally, the goal of meta-learning is to train a model on a variety of learning tasks such that it can generalize to new learning tasks. There are mainly three series of meta-learning methods: 
(1) black-box methods train black-box meta learners (e.g., neural networks) to output model parameters for tasks by standard supervised learning \cite{mann, metanet, snail}; 
(2) optimization-based methods train the model’s initial parameters such that the model has maximal performance on a new task after the parameters have been updated through one or more gradient steps computed with the data from that new task \cite{maml, ARML}; 
(3) metric-based methods, aka non-parametric methods, learn universal matching metrics which can generalize among all tasks for classification \cite{siamese, prototypical, matching}. 

Some recent works consider tackling problems related to knowledge graphs by meta-learning. For example, as for few-shot link prediction in knowledge graphs, which requires predicting triples with only observing a few samples for a specific relation, GMatching \cite{GMatching} and subsequent metric-based methods \cite{FSRL, FAAN} learn matching metrics by graph-structures and learned embeddings; MetaR \cite{MetaR} and GANA \cite{GANA} leverage optimization-based methods for fast adaption of relation embeddings. 
Furthermore, GEN \cite{GEN} tackles the few-shot out-of-graph link prediction problem by a meta-learning framework that meta-learns embeddings for few-shot new entities which connected to the original KG. 
Furthermore, in the area of graph representation learning, L2P-GNN \cite{L2P-GNN} is proposed for pre-training GNNs by meta-learning to make GNNs handle new graphs for downstream tasks, and MI-GNN \cite{MI-GNN} customizes the inductive model to each graph under a meta-learning paradigm to achieve inductive node classification across graphs.

Existing meta-learning works for knowledge graphs mainly focus on applying meta-learning to few-shot scenarios. 
In contrast, in our work, we aim to use meta-learning to simulate the tasks of embedding unseen entities, which makes our model generalize to KGs with unseen entities in the inductive setting.

\section{Problem Formulation}
\label{sec:problem-formulation}

A knowledge graph is defined as $\mathcal{G} = ( \mathcal{E}, \mathcal{R}, \mathcal{P})$, where $\mathcal{E}$ is a set of entities and $\mathcal{R}$ is a set of relations. $\mathcal{P} = \{ (h, r, t) \} \subseteq \mathcal{E} \times \mathcal{R} \times \mathcal{E}$ is a set of triples. 
Conventional KGE methods train an entity embedding matrix $\mathbf{E} \in \mathbb{R}^{|\mathcal{E}| \times d}$ and a relation embedding matrix $\mathbf{R} \in \mathbb{R}^{|\mathcal{R}| \times d}$ for a fixed set of $\mathcal{E}$ and $\mathcal{R}$, where such embeddings should follow the assumption from the score function $s$ related to a specific KGE method, namely, the embeddings should satisfy \textit{reasonability}. 
More precisely, $s(h, r, t)$ for $(h,r,t) \in \mathcal{P}$ should be higher than $s(h^{\prime}, r^{\prime}, t^{\prime})$ for $(h^{\prime},r^{\prime},t^{\prime}) \notin \mathcal{P}$, and $s(h, r, t)$ is calculated by corresponding embeddings based on a specific KGE method. Such reasonable embeddings can be used for various in-KG and out-KG tasks.

We next give the definition for the inductive knowledge graph embedding.
Formally, given a set of source KGs $\mathcal{G}_S = \{\mathcal{G}^{(i)}_s $ $= ( \mathcal{E}^{(i)}_s, \mathcal{R}^{(i)}_s, \mathcal{P}^{(i)}_s)\}_{i=1}^{n_s}$ and a set of target KGs consisting of entities unseen in source KGs,
\begin{equation}
\begin{aligned}
    &\mathcal{G}_T = \{\mathcal{G}^{(i)}_t = ( \mathcal{E}^{(i)}_t, \mathcal{R}^{(i)}_t, \mathcal{P}^{(i)}_t)\}_{i=1}^{n_t} \\
    &\text{s.t.} (\cup_{i=1}^{n_t} \mathcal{E}^{(i)}_t) \cap (\cup_{i=1}^{n_s} \mathcal{E}^{(i)}_s) = \emptyset, (\cup_{i=1}^{n_t} \mathcal{R}^{(i)}_t) \subseteq (\cup_{i=1}^{n_s} \mathcal{R}^{(i)}_s).
\label{eq:target-kg}
\end{aligned}
\end{equation}
The goal of inductive knowledge graph embedding is to learn a function $f$ on 
source KGs $\mathcal{G}_S$, where the $f$ can map entities in $\mathcal{G}_S$ into embeddings with the property of \textit{reasonability}, as introduced above, and is able to generalize to target KGs $\mathcal{G}_T$.
Such reasonable embeddings can be used for various in-KG and out-of-KG tasks on target KGs, and achieve inductive knowledge graph embedding.

Note that the number of source KGs and target KGs won't affect the training and applying of $f$, at least for the method in our paper.
Thus for the simplicity of notations, we will consider the scenario of one source KG and one target KG for describing our proposed framework, which means $n_s = n_t = 1$, $\mathcal{G}_S = ( \mathcal{E}_s, \mathcal{R}_s, \mathcal{P}_s)$  and $\mathcal{G}_T = ( \mathcal{E}_t, \mathcal{R}_t, \mathcal{P}_t)$. 

\section{methodology}

Conceptually, we focus on training an embedding-based model to capture meta-knowledge (i.e., knowledge about transferable structural patterns) in the source KG, and transferring that for producing reasonable embeddings for the target KG to benefit various tasks. 
Such meta-knowledge is implicitly captured by the function $f$ aforementioned, and can be used for producing entity embeddings. To achieve meta-knowledge transfer, our \model~needs to solve the following three sub-problems:
\begin{itemize}[leftmargin=14pt]
\item[\textbf{P1}] How to model the meta-knowledge?
\item[\textbf{P2}] How to learn the meta-knowledge in the source KG $\mathcal{G}_S$?
\item[\textbf{P3}] How to adapt the meta-knowledge to the target KG $\mathcal{G}_T$?
\end{itemize}

\begin{figure*}[t]
\centering
\includegraphics[scale=0.45]{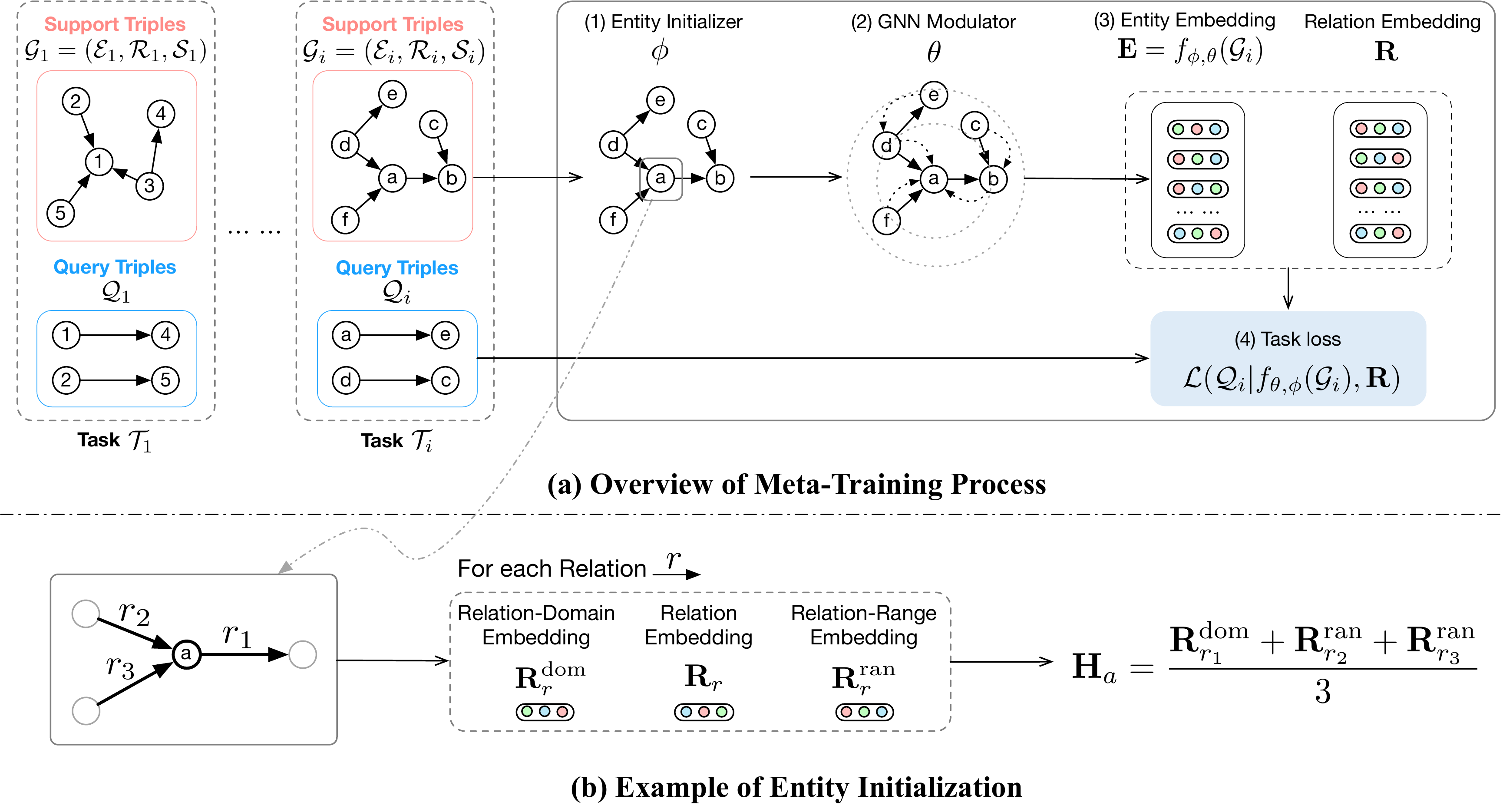}
\caption{The illustration of the overall meta-training process.}
\label{fig:method}
\end{figure*}

\subsection{Modeling of Meta-Knowledge}
\label{sec:modeling-of-meta-knowledge}

In this part, we solve the sub-problem \textbf{P1: How to model the meta-knowledge}. 
We focus on designing modules that can produce entity embeddings based on its neighbor structural information, simulating the process of cognizing a new entity for humans.
For an entity, its most natural structural information can be indicated by relations around it. Thus, first we design an \textit{Entity Initializer} to initialize the embedding of each entity using the information of relations connected to it. 
However, such initialized entity embeddings are naive, as they only convey type-level information but not instance-level information. For example, if two entities are both the head of relation \texttt{student\_of}, we can only infer that they are two students but not exactly who they are.
To solve this problem, we introduce a \textit{GNN Modulator} to modulate the initialized embedding for each entity based on its multi-hop neighborhood structure.
These two modules are both biased toward graph symmetries. Namely, entities with similar connected relations and multi-hop neighborhoods will get similar embeddings through these two modules. Thus they are capable of capturing the transferable structural patterns in KGs.
In the following, we describe the process of the entity initializer and the GNN modulator given a KG instance $\mathcal{G} = (\mathcal{E}, \mathcal{R}, \mathcal{P})$.


\subsubsection{Entity Initializer}
This module is designed for capturing the type-level information of entities. Thus, besides the conventional relation embedding matrix $\mathbf{R} \in \mathbb{R}^{n_r \times d}$, we design a learnable relation-domain embedding matrix $\mathbf{R}^{\mathrm{dom}} \in \mathbb{R}^{n_r \times d}$ and a learnable relation-range embedding matrix $\mathbf{R}^{\mathrm{ran}} \in \mathbb{R}^{n_r \times d}$ to represent the implicit type features of head and tail entities for each relation, where $n_r$ is the number of relations ($n_r=|\mathcal{R}_s|$ when we train this model on ${\mathcal{G}_S}$) and $d$ is the dimension for embeddings. 
Specifically, $\mathbf{R}$ reserving the internal semantics of relations is used for learning relation embedding in Section \ref{sec:trainig}. $\mathbf{R}^{\mathrm{dom}}$ and $\mathbf{R}^{\mathrm{ran}}$ containing information about implicit type of entities is used for entity initialization.
For a specific relation $r$, its relation embedding, relation-domain embedding and relation-range embedding are represented as $\mathbf{R}_{r}$, $\mathbf{R}_{r}^{\mathrm{dom}}$ and $\mathbf{R}_{r}^{\mathrm{ran}}$ respectively, as shown in Figure \ref{fig:method} (b).

For the KG instance $\mathcal{G} = (\mathcal{E}, \mathcal{R}, \mathcal{P})$, \model~initializes the entity embeddings based on their connected relations. Specifically, for an entity $e \in \mathcal{E}$, its initialized embedding $\mathbf{H}_{e}$ is calculated by the average of the 
relation-range and relation-domain embedding of its ingoing and outgoing relations:
\begin{equation}
	\mathbf{H}_{e} = \frac{\sum_{r \in \mathcal{O}(e)}{\mathbf{R}^{\mathrm{dom}}_{r}} + \sum_{r \in \mathcal{I}(e)}{\mathbf{R}^{\mathrm{ran}}_{r}}}{|\mathcal{O}(e)|+|\mathcal{I}(e)|},
\label{eq:ent-init}
\end{equation}
where $\mathcal{I}(e) = \{r|\exists x,(x,r,e) \in \mathcal{P}\}$
denotes the ingoing relation set for entity $e$, $\mathcal{O}(e) = \{r|\exists x,(e,r,x) \in \mathcal{P}\}$ denotes the outgoing relation set for entity $e$. A visual illustration of using $\mathbf{R}^{\mathrm{dom}}$ and $\mathbf{R}^{\mathrm{ran}}$ for entity initialization is shown in Figure \ref{fig:method} (b).

\subsubsection{GNN Modulator}

This module is designed to capture entities' instance-level information via structure information from their multi-hop neighborhoods. So far, several previous works have shown that the graph neural network has the capability of capturing the local structure information of knowledge graphs \cite{RGCN, GraIL}. Thus, the modulation of initialized entity embeddings is archived by a GNN modulator. Following R-GCN \cite{RGCN}, our GNN modulator calculates the update in the $l$-th layer for an entity $e$ as follows:
\begin{equation}
	\mathbf{h}_{e}^{l} = \alpha \left( \frac{1}{|\mathcal{N}(e)|} \sum_{(h,r)\in \mathcal{N}(e)} \mathbf{W}_r^{l} \mathbf{h}_{h}^{l-1} + \mathbf{W}_0^{l} \mathbf{h}_{e}^{l-1} \right), 
\label{eq:ent-gnn-update}
\end{equation}
where $\alpha$ is an activation function and we use ReLU here; $\mathcal{N}(e) = \{(h,r)|(h,r,e) \in \mathcal{P}\}$ denotes 
the set of head entity and relation pair of immediate ingoing neighbor triples of entity $e$;
$\mathbf{W}_r^l$ is the relation-specific transformation matrix for relation $r$ in the $l$-th layer; $\mathbf{W}_0^l$ is a self-loop transformation matrix for entities
in the $l$-th layer; $\mathbf{h}_e^{l}$ denotes the hidden entity representation of entity $e$, and the input representation $\mathbf{h}_e^0 = \mathbf{H}_e$.

To make full use of hidden entity representations of every layer and let the model leverage the most appropriate neighborhood range flexibly for each entity, we apply a jumping knowledge (JK) structure \cite{JK-Net} based on the concatenation of hidden representations, as follows:
\begin{equation}
	\mathbf{E}_{e} = \mathbf{W}^{\mathrm{JK}} \bigoplus_{l=0}^{L} \mathbf{h}_{e}^{l},
\label{eq:jk-connect}
\end{equation}
where $\mathbf{E}_e$ is the final entity embedding for the entity $e$; $\bigoplus$ denotes successive concatenation; $L$ is the number of GNN modulator layers; $\mathbf{W}^{\mathrm{JK}}$ is a matrix to transform concatenated hidden representations to entity embeddings.

\subsubsection{Summary} Given a KG $\mathcal{G} = (\mathcal{E}, \mathcal{R}, \mathcal{P})$, the process of entity initializer is,
\begin{equation}
\mathbf{H} = \text{INIT}_{\theta}(\mathcal{G}),
\label{eq:INIT}
\end{equation}
where $\mathbf{H}$ is the initialized entity embeddings of $\mathcal{E}$, and $\theta$ is the parameter set including $\mathbf{R}^{\mathrm{dom}}$ and $\mathbf{R}^{\mathrm{ran}}$.
Next, the process of the GNN modulator can be viewed as:
\begin{equation}
\mathbf{E} = \text{MODULATE}_{\phi}(\mathcal{G}, \mathbf{H}),
\label{eq:MODULATE}
\end{equation}
where $\mathbf{E}$ is the output entity embeddings of $\mathcal{E}$, and $\phi$ is the parameter set of the GNN modulator. The whole procedure can be viewed as:
\begin{equation}
\mathbf{E} = f_{\theta,\phi}(\mathcal{G}) = \text{MODULATE}_{\phi}(\mathcal{G}, \text{INIT}_{\theta}(\mathcal{G})).
\label{eq:REP}
\end{equation}
The function $f_{\theta,\phi}$ can map entities in the KG into embeddings as mentioned in Section \ref{sec:problem-formulation}. We use this function and its parameters to model transferable meta-knowledge, and we describe how to train this function for outputting reasonable entity embeddings in the next part.


\subsection{Learning of Meta-Knowledge}

In this part, we solve the sub-problem \textbf{P2: how to learn the meta-knowledge in the source KG $\mathcal{G}_S$}, via meta-learning,
namely ``learning to learn''. 
In this part, we first illustrate the concept of meta-learning and the corresponding setting in \model. Then, we describe the training regime based on meta-learning.

\subsubsection{Meta-Learning Setting}
\label{sec:meta-learning-setting}

The goal of training is to make the above $f_{\theta,\phi}$ enable to output reasonable entity embeddings given any KG with unseen entities, namely learning to produce entity embeddings. 
Inspired by the concept ``learning to learn'' of meta-learning \cite{maml}, we train \model~on a set of tasks.
Specifically, we sample a set of sub-KGs from the source KG $\mathcal{G}_S$, and treat the entities in such sub-KGs as unseen ones for simulating target KGs in inductive settings. 
Furthermore, to formulate a task for each sub-KG, we split a part of triples as query triples, and treat the remaining triples as the support triples. 
The support triples are used to produce entity embeddings, and the query triples are used to evaluate the reasonability of produced embeddings and calculate training loss. Formally, a task $\mathcal{T}_{i}$ is defined as follows:
\begin{equation}
\begin{aligned}
    \mathcal{T}_{i} = ( &\mathcal{G}_i = 
    \left( \mathcal{E}_i, \mathcal{R}_i, \mathcal{S}_i \right), 
    \mathcal{Q}_i ) \\
    \text{s.t.} \quad 
    &\mathcal{E}_i \cap \mathcal{E}_s = \emptyset, \mathcal{R}_i \subseteq \mathcal{R}_s,
\label{eq:task}
\end{aligned}
\end{equation}
where $\mathcal{S}_{i}$ is the support triple set and $\mathcal{Q}_{i}$ is the query triple set $\mathcal{E}_s$ and $\mathcal{R}_s$ are entity and relation set of the source KG $\mathcal{G}_s$. For simulating the inductive setting of target KGs (Equation \ref{eq:target-kg}), we treat entities in each task as unseen, and this can be implemented by re-labeling entities as described in Section \ref{sec:task-sampling}.
Since these tasks simulate the scenario of the target KG with unseen entities, our model meta-trained on these tasks can naturally generalize to the target KG.

\subsubsection{Meta-Training Regime}
\label{sec:trainig}

Following the meta-learning setting, meta-knowledge is learned based on the tasks sampled from $\mathcal{G}_S$, which is also referred to as the meta-training process. During meta-training, for each task $\mathcal{T}_i = (\mathcal{G}_i=(\mathcal{E}_i, \mathcal{R}_i, \mathcal{S}_i), \mathcal{Q}_i)$, the entity embeddings are obtained based on its support triples and evaluated by its query triples. 
Formally, the overall meta-training objective is,
\begin{equation}
	\min_{\theta, \phi, \mathbf{R}} \sum_{i=1}^{m} \mathcal{L}_i = \min_{\theta, \phi, \mathbf{R}} \sum_{i=1}^{m} \mathcal{L} (\mathcal{Q}_i | f_{\theta, \phi}(\mathcal{G}_i), \mathbf{R}),
\label{eq:overall-objective}
\end{equation}
where $\mathbf{R} \in \mathbb{R}^{|\mathcal{R}_s| \times d}$ is the relation embedding matrix for all relations, and $f_{\theta, \phi}(\mathcal{G}_i)$ is used for outputting entity embedding matrix for entities in current task (i.e., $\mathcal{E}_i$); $\{\theta, \phi, \mathbf{R} \}$ are learnable parameters; $m$ denotes the number of all sampled tasks.
The visual illustration of the meta-training process is shown in Figure \ref{fig:method} (a). 

The calculation of training loss is dependent on different score functions of different KGE methods. For example, as for task $\mathcal{T}_i$, the score function based on TransE \cite{TransE} is shown as follows:
\begin{equation}
\begin{aligned}
	s(h, r, t) &= - \|\mathbf{E}_h + \mathbf{R}_r - \mathbf{E}_t\|,
\label{eq:score-function}
\end{aligned}
\end{equation}
where $\mathbf{E} = f_{\theta, \phi}(\mathcal{G}_i)$; $\| \cdot \|$ denotes the L2 norm. Many conventional KGE methods can be used in our model. We use four representative methods, including TransE \cite{TransE}, DistMult \cite{DistMult}, ComplEx \cite{ComplEx} and RotatE \cite{RotatE} in our paper, and the details of their score functions can be found in Table \ref{tab:score-function}.

\begin{table}[t]
\centering
\caption{Score function $s(h, r, t)$ of typical knowledge graph models. $\mathbf{h}, \mathbf{r}, \mathbf{t}$ are embeddings correspond to $h, r, t$. $\operatorname{Re(\cdot)}$ denotes the real vector component of a complex valued vector. $\circ$ denotes the Hadamard product.}
\begin{tabular}{lcc}
\toprule
Model & Score Function & Vector Space \\
\midrule
TransE & $-|| \mathbf{h} + \mathbf{r} - \mathbf{t} ||$ & $ \mathbf{h}, \mathbf{r}, \mathbf{t} \in \mathbb{R}^d $ \\
DistMult & $\mathbf{h}^\top \operatorname{diag}(\mathbf{r}) \mathbf{t}$ & $ \mathbf{h}, \mathbf{r}, \mathbf{t} \in \mathbb{R}^d $ \\
ComplEx & $\operatorname{Re}(\mathbf{h}^\top \operatorname{diag}(\mathbf{r}) \overline{\mathbf{t}})$ & $ \mathbf{h}, \mathbf{r}, \mathbf{t} \in \mathbb{C}^d $ \\
RotatE & $-|| \mathbf{h} \circ \mathbf{r} - \mathbf{t} ||$ & $ \mathbf{h}, \mathbf{r}, \mathbf{t} \in \mathbb{C}^d $ \\
\bottomrule
\end{tabular}
\label{tab:score-function}
\end{table}

During training, we apply the loss function based on self-adversarial negative sampling \cite{RotatE} on query triples $\mathcal{Q}_i$ for each task, and the details are described as follows: 
\begin{equation}
\begin{aligned}
\mathcal{L}_i = &\sum_{(h, r, t) \in \mathcal{Q}_i} - \log \sigma \left( \gamma + s(h,r,t) \right) \\
&-\sum_{i=1}^{k} p\left(h_i^{\prime}, r, t_{i}^{\prime}\right) \log \sigma\left(- \gamma - s(h_i^{\prime}, r, t_{i}^{\prime}) \right),
\label{eq:loss-function}
\end{aligned}
\end{equation}
where $\sigma$ is the sigmoid function; $\gamma$ is a fixed margin; $k$ is the number of negative samples for each triple; $(h_i^{\prime}, r, t_i^{\prime})$ is the $i$-th negative triple by corrupting head or tail entity; $p(h_i^{\prime}, r, t_{i}^{\prime})$ is the self-adversarial weight calculated by,
\begin{equation}
    p\left(h_j^{\prime}, r, t_{j}^{\prime}\right) = \frac{\exp \beta s(h_j^{\prime}, r, t_{j}^{\prime})}{\sum_{i} \exp \beta s(h_i^{\prime}, r, t_{i}^{\prime})}, 
\label{eq:adv-sample}
\end{equation}
where $\beta$ is the temperature of sampling.

The learning of meta-knowledge is meta-training the modules described in Section \ref{sec:modeling-of-meta-knowledge} on the source KG $\mathcal{G}_S$ by meta-learning described in this section. After meta-training, the function $f_{\theta, \phi}$ can be used for producing entity embeddings for tasks with unseen entities and is able to generalize to the target KG $\mathcal{G}_T$ to achieve inductive knowledge graph embedding. The procedure of producing entity embeddings on $\mathcal{G}_T$ and servicing downstream tasks is meta-knowledge adaption, and we describe it in the following.

\subsection{Adaption of Meta-Knowledge}
\label{sec:adapt}
In this part, we tackle the sub-problem \textbf{P3: how to adapt the meta-knowledge to the target KG $\mathcal{G}_T$}.
Adapting to a given target KG $\mathcal{G}_T = (\mathcal{E}_t, \mathcal{R}_t, \mathcal{P}_t)$ is a reflection of meta-knowledge transfer, and this step is straightforward as the ability to output entity embeddings from $f_{\theta, \phi}(\mathcal{G}_T)$ allows \model~to adapt to various in-KG and out-of-KG tasks flexibly. \model~offers two meta-knowledge adaption regimes as follows.

\subsubsection{Freezing} In this adaption mode, we freeze the parameters of the meta-trained function $f_{\theta, \phi}$, and relation embedding matrix $\mathbf{R}$. Then we treat \model~as an entity embedding producer, and use such entity embeddings produced on target KGs to finish downstream tasks. 
For example, in link prediction, such embeddings can be used for conducting link prediction directly and achieving inductive link prediction.
In question answering, such embeddings can be used as features for KGs related to QA pairs, and SOTA QA models can be trained on top of the produced embeddings.

\subsubsection{Fine-tuning} 

In this adaption mode, the components of meta-trained \model~including $f_{\theta, \phi}$ and $\mathbf{R}$ can be trained based on specific tasks. 
For example, in link prediction, a meta-trained \model~can be fine-tuned based on triples in target KGs.

\section{Experiments}

In this section, we evaluate \model~on two representative KG-related tasks, an in-KG task---link 
prediction, and an out-of-KG task---question answering, in inductive settings. 
We conduct extensive experiments to show the effectiveness of our method, and the following are key questions we explore during experiments:
\begin{itemize}[leftmargin=16pt]
    \item[\textbf{Q1}] How does the performance of \model~compare to baselines in conducting link prediction for KGs in inductive settings?
    \item[\textbf{Q2}] How does the performance of \model~compare to baselines in conducting question answering for KGs in inductive settings?
    \item[\textbf{Q3}] 
    How much do the two modules for modeling meta-knowledge and the meta-learning setting contribute?
    What is the influence of target KG's sparsity on \model?
\end{itemize}
We first introduce the training setups for both tasks in Section \ref{sec:training-setup}, and then describe the details and results of experiments on two tasks in Section \ref{sec:exp-link-prediction} and \ref{sec:exp-qa}. Finally, we give some further analysis in Section \ref{sec:model-analysis}.

\subsection{Training Setup}
\label{sec:training-setup}

In this part, we describe the overall training settings, including the strategy of sampling tasks from a source KG for meta-learning and implementation details of training \model.

\subsubsection{Task Sampling}
\label{sec:task-sampling}

As described in Section \ref{sec:meta-learning-setting}, the tasks for meta-training are sampled from a source KG $\mathcal{G}_s = (\mathcal{E}_s, \mathcal{R}_s, \mathcal{P}_s)$.
During experiment, we sample each task $\mathcal{T}_i$ with following steps:
\begin{itemize}[leftmargin=16pt]
    \item[(1)] \textit{Random Walk Sampling}. First, we sample an entity $e \in \mathcal{E}_{s}$, from which we conduct $n_{rw}$ times random walks with length $l_{rw}$, resulting in a set of entities $\mathcal{E}_{rw}$.
    Second, we sample an entity $e^\prime \in \mathcal{E}_{rw}$, and repeat the first step with getting a new entity set $\mathcal{E}_{rw}^\prime$, and update $\mathcal{E}_{rw}$ that $\mathcal{E}_{rw} := \mathcal{E}_{rw} \cup \mathcal{E}_{rw}^\prime$.
    \item[(2)] \textit{Sub-KG Induction.}  We repeat the second step in (1) with $t_{rw}$ times, and induce the sub-KG triples $\{(h,r,t)| h \in \mathcal{E}_{rw}, t \in \mathcal{E}_{rw}, (h,r,t) \in \mathcal{P}_s\}$ as the triples for the task $\mathcal{T}_i$.
    \item[(3)] \textit{Re-labeling.} Finally, we anonymize entities in $\mathcal{T}_i$ by re-labeling them to be $\{1,2,\cdots,|\mathcal{E}_{rw}|\}$ in an arbitrary order, 
    as they are unseen entities.
\end{itemize}
We randomly split a part of triples in this sub-KG as query triples $\mathcal{Q}_i$, and the remaining ones are support triples $\mathcal{S}_i$, as described in the Equation \ref{eq:task}.

\subsubsection{Implementation Details}

Our model is implemented in PyTorch \cite{pytorch} and DGL \cite{DGL}, and the experiments are conducted on RTX 3090 GPUs with 24GB RAM. 
For the inductive link prediction/question answering task, we employ a 3/2-layer GNN modulator and the dimension of embeddings and latent representations in the GNN modulator are 32/200. 
We also adopt the basis-decomposition \citet{RGCN} as a regularization on the GNN modulator, and the number of bases is 4. Note that all these settings are set for fair comparison with baselines.
Moreover, for more hyper-parameters in link prediction and question answering tasks, we set the batch size as 64 and 16 respectively, the learning rate as 0.01, the number of training epochs as 10 and 5 respectively, the fixed margin $\gamma$ as 10, the number of negative samples $k$ as 32, the temperature of sampling $\beta$ as 1, the number of training tasks as 10,000, the number of validation tasks as 200, the random walk parameters $n_{rw}=10$, $l_{rw}=5$ and $t_{rw}=10$.
Furthermore, for each experiment related to the link prediction task, we run 5 times with average results reported, and run 3 times for the question answering task.


\input{table-results-lp}

\subsection{In-KG Task: Link Prediction}
\label{sec:exp-link-prediction}

The link prediction task for KGs in the inductive setting has been explored in some previous work \cite{GraIL,compile}. We follow their proposed benchmarks and baselines to show the effectiveness of our model. The overall procedure of inductive link prediction is first training a model on the source KG and then testing it on the target KG consisting of entities unseen during training by predicting missing triples \cite{GraIL}.

\subsubsection{Datasets and Evaluation Metrics}
We use datasets derived from original WN18RR \cite{ConvE}, FB15k-237 \cite{fb15k237} and NELL-995 \cite{DeepPath} by \citet{GraIL}, which are created for inductive link prediction.
Each dataset is sampled as four different versions depending on various scales for robust evaluation.
Specifically, each dataset has two disjoint parts sampled from original benchmarks, corresponding to the source KG and the target KG in the inductive setting.
The sampling details can be found in \citet{GraIL}.
The statistics of datasets are shown in Table \ref{tab:statistic-lp}.

We report Mean Reciprocal Rank (MRR) and Hits at N (Hits@N) \cite{olddog} to evaluate the performance of link prediction in target KGs. The results are averaged by head and tail prediction. 
Following baselines \cite{GraIL, compile}, the link prediction evaluation results are approximated by ranking each test triple among 50 other randomly sampled candidate negative triples five times. 

\subsubsection{Baselines}
We compare our \model~with two kinds of methods, rule-learning-based and GraIL-based methods. Rule-learning-based methods consist of RuleN \cite{rulen}, which explicitly extracts rules from KGs, and Neural-LP \cite{neural-lp} and DRUM \cite{drum}, which learn rules in an end-to-end differentiable manner. GraIL-based methods consist of GraIL \cite{GraIL}, which learns a GNN to conduct relation prediction inductively and can handle unseen entities, and its improvement method, CoMPILE \cite{compile}.

\subsubsection{Adaption Details}
As described in Section \ref{sec:adapt}, for adapting \model~to the link prediction task, we can freeze its parameters, and use it to produce entity embeddings on target KGs to conduct inductive link prediction directly. This experimental setup is the same as the setup of prior models achieving inductive link prediction, and we show results for \model~on freezing adaption mode in Table \ref{tab:link-prediction-rst}.
Furthermore, unlike baselines, our model provides fine-tuning adaption that trains \model~on triples in target KGs, and we calculate the loss based on the same loss function as Equation (\ref{eq:loss-function}) for fine-tuning optimization.
During fine-tuning, we set the batch size as 512, the learning rate as 0.001, and $\gamma=10$, $k=64$, $\beta=1$.

\begin{figure*}[t]
\centering
\includegraphics[scale=0.53]{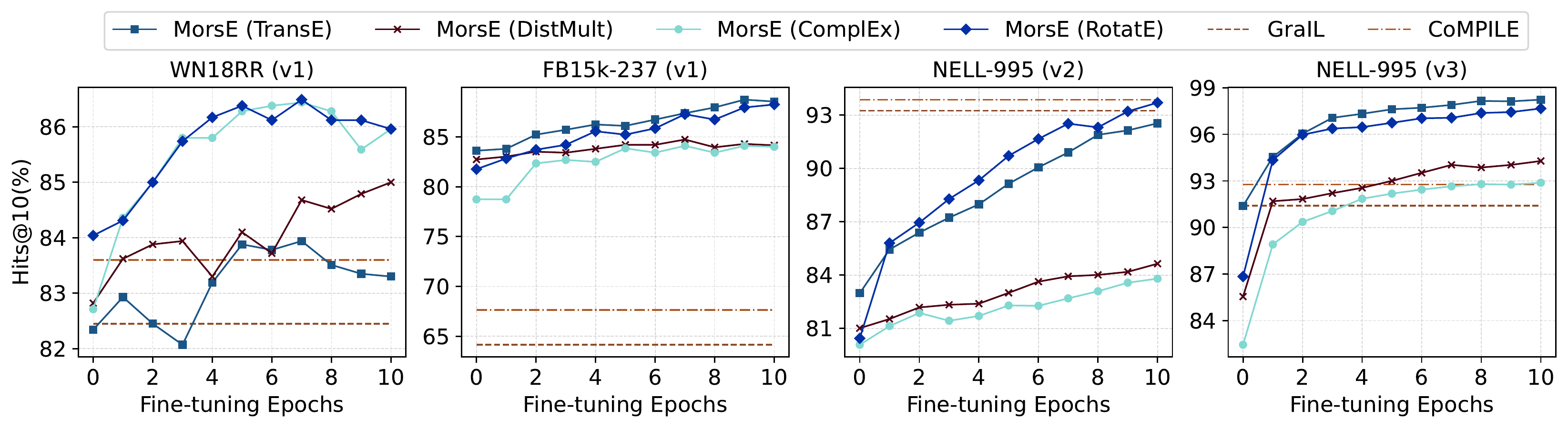}
\caption{Results with different fine-tuning epochs.}
\label{fig:finetune}
\end{figure*}

\subsubsection{Result Analysis}

In Table \ref{tab:link-prediction-rst}, we show the results of inductive link prediction. For a fair comparison, results of \model~are from freezing adaption.
Note that our model is an embedding-based framework that can be equipped with different score functions from conventional KGE methods, and thus results of \model~equipped with four typical KGE methods, TransE, DistMult, ComplEx, and RotatE, are reported.
From Table \ref{tab:link-prediction-rst}, we can see that \model~outperforms baselines on most datasets and achieves the state-of-the-art performance in inductive link prediction. 
Even though we only highlight the best results, \model~with different KGE methods consistently outperforms baselines. For example, among different versions of FB15k-237, the results of \model~with TransE, DistMult, ComplEx and RotatE, increase on average, by 14.6\%, 13.0\%, 10.8\%, 13.7\% relatively. More importantly, as a framework of inductive knowledge graph embedding, \model~could continuously take advantage of the development in the area of knowledge graph embedding, while other baselines can not.

Moreover, compared with baselines which are directly adapted to a target KG after being trained on a source KG, \model~gives a more flexible choice for adaption---fine-tuning on the target KG. In Figure \ref{fig:finetune}, we show the results of \model~with different fine-tuning epochs on various datasets. We find that after fine-tuning just a few epochs, \model~obtains significant improvements for different KGE methods. Such fine-tuning is efficient and effective. For example, in NELL-995 (v3), even though the results of \model~with TransE and RotatE are worse than baselines in Table~\ref{tab:link-prediction-rst}, just 1 quick epoch fine-tuning will make \model~outperform CoMPILE and achieve the state-of-the-art performance on NELL-995 (v3) (the fourth line chart in Figure \ref{fig:finetune}).

Overall, we show that our proposed \model~obtains better performance compared with baselines on inductive link prediction tasks, indicating that \model~can achieve inductive knowledge graph embedding effectively (\textbf{Q1}). 
Furthermore, besides adapting the model directly using the freezing setting to a target KG like conventional inductive KGC methods, our proposed \model~can be easily fine-tuned on the target KG, and such fine-tuning is efficient and effective for further performance improvement.



\subsection{Out-of-KG Task: Question Answering}
\label{sec:exp-qa}

To show that our model can achieve inductive knowledge graph embedding, which can solve various KG-related tasks, including out-of-KG tasks, we evaluate \model~on answering questions using knowledge from KGs \cite{QA-GNN}.
Unlike the inductive link prediction task, QA with KGs in inductive settings has not been studied, so we create an inductive version based on conventional QA tasks. We first describe this problem statement in the following.

\subsubsection{Problem Statement}

This problem focuses on answering natural language questions using knowledge from a pre-trained language model and a structured KG, and more details can be found in \citet{QA-GNN}. We mainly describe the inductive setting for this QA problem here.
Given a question $q$ and an answer choice $a$ (a QA pair), prior works \cite{kagnet, QA-GNN} will find the entities that appear in either the question or answer choice and treat them as topic entities $\mathcal{V}_{q,a}$, and then extract a sub-KG $\mathcal{G}_{sub}^{q,a}$ from a background KG $\mathcal{G}_{bg}$ based on $\mathcal{V}_{q,a}$. $\mathcal{G}_{sub}^{q,a}$ comprises all entities on the $k$-hop paths between entities in $\mathcal{V}_{q,a}$. 
To construct KGs in the inductive setting, after extracting sub-KGs for all the QA pairs, we remove entities together with their related triples in the sub-KGs from the background KG $\mathcal{G}_{bg}$, and denote the remaining part as $\mathcal{G}_{bg}^{\prime}$. 
For the implementation of \model, we treat $\mathcal{G}_{bg}^{\prime}$ as a source KG, and the sub-KGs
for QA pairs as target KGs. The entities in the target KGs are not seen in the source KG. 


\subsubsection{Datasets and Evaluation Metrics}
We use the dataset CommonsenseQA \cite{CommonsenseQA}, which is a 5-way multiple choice QA task that requires reasoning with commonsense knowledge and contains 12,102 questions.
For each QA
pair, its corresponding commonsense is a sub-graph extracted from \textit{ConceptNet} \cite{conceptnet}, which is a general-domain knowledge graph.
As mentioned above, we remove the entities appearing in KGs of QA pairs from \textit{ConceptNet}, and then treat the remaining part as the source KG and KGs of QA pairs as the target KGs. 
Finally, there are 17 relations, 650,763 entities and 1,148,711 triples in the source KG ($\mathcal{G}_{bg}'$). 
Following prior works, we conduct the experiments on the in-house data split used in \citet{kagnet}, and report the accuracy (Acc.) of the development set (IHdev) and test set (IHtest).

\subsubsection{Baselines}
Following prior works on question answering with language models and knowledge graphs \cite{kagnet,MHGRN,QA-GNN}, we first compare \model~with a vanilla fine-tuned language model---RoBERTa-large \cite{roberta}, which does not use the KG. 
Furthermore, we compare \model~with QA-GNN \cite{QA-GNN}, and its baselines, including RN \cite{RN}, RGCN \cite{RGCN}, GconAttn \cite{GconAttn}, KagNet \cite{kagnet}, MHGRN \cite{MHGRN}. RN, RGCN and GconAttn are relation-aware graph neural networks for KGs, and KagNet and MHGRN are methods model paths in KGs for QA pairs.
For a fair comparison, we use the same language model in all the baselines and our model.

\subsubsection{Adaption Details}
\label{sec:qa-adaption-details}
According to Section \ref{sec:adapt}, we adapt \model~to the question answering task by freezing mode and the inductive knowledge graph embeddings produced by \model~are used as input entity features in QA-GNN \cite{QA-GNN}.
We use meta-trained \model~to obtain entity embeddings for $\mathcal{G}_{sub}^{q,a}$ for each QA pair $(q, a)$ and train QA-GNN \cite{QA-GNN} on top of the produced embeddings.

In general, given a QA pair $(q, a)$ and a 
KG $\mathcal{G}^{q,a}_{sub}$ related to this QA pair, the goal of question answering is to calculate the probability of $p(a | q)$ based on the QA context and $\mathcal{G}^{q,a}_{sub}$.
In QA-GNN, this probability is calculated by,
\begin{equation}
p(a|q) \propto \exp(\mathbf{MLP}(\boldsymbol{z}^{\mathrm{LM}}, \text{QA-GNN}(q, a, \mathcal{G}^{q,a}_{sub}))),
\end{equation}
where $\boldsymbol{z}^{\mathrm{LM}}$ is the output representation of the QA pair from a language model (LM). QA-GNN is conducted on a KG where $q$ and $a$ are connected to $\mathcal{G}^{q,a}_{sub}$. During conducting QA-GNN, the input feature for $q$ and $a$ are their representations from the LM, and the input feature $\mathbf{E}^{\mathrm{in}}$ for entities in $\mathcal{G}^{q,a}_{sub}$ are obtained by their text information in \textit{ConceptNet} and pre-trained \textsc{Bert-Large}, which is proposed in \citet{MHGRN}.

To adapt \model~to QA-GNN, we fuse the entity embedding from meta-trained \model~with input entity features in QA-GNN,
\begin{equation}
\begin{aligned}
\mathbf{E} &= f_{\theta, \phi}(\mathcal{G}^{q,a}_{sub}), \\
\mathbf{E}^{\mathrm{in}} :&= \mathbf{W}^{\mathrm{f}}\left[\mathbf{E}; \mathbf{E}^{\mathrm{in}}\right]. \\
\end{aligned}
\end{equation}
During adapting \model~to QA-GNN (i.e., QA-GNN+\model), we train it using the same hyper-parameter settings as original QA-GNN, and use the trained QA-GNN parameters as the initialization. The learnable parameters include QA-GNN parameters and the $\mathbf{W}^{\mathrm{f}}$ for embedding fusion, and the $\theta$ and $\phi$ are fixed.

\subsubsection{Result Analysis}
Results are shown in Table \ref{tab:qa}, where \model~is equipped with the general and effective KGE method TransE.
Table \ref{tab:qa} shows that our proposed \model~achieves the best performance, and obtains consistent improvement compared to the baselines. In comparison with the best results among baselines, the results of \model~increase by 1.5\% and 2.9\% relatively on the development and test set. For each QA-related KG, \model~produces entity embeddings for it directly using freezing adaption, since the task training objective is QA and the freezing adaption will make the model more focused on learning parameters related to QA tasks.
Based on the state-of-the-art QA results in Table~\ref{tab:qa}, we could conclude that \model~successfully achieves
inductive knowledge graph embedding, which helps question answering based on KGs in the inductive setting (\textbf{Q2}).

\begin{table}[t]
\renewcommand\arraystretch{1.1}
\centering
\caption{Performance (\%) for question answering on CommonsenseQA. Results of baselines are taken from \citet{QA-GNN}.}
\begin{tabular}{lcc}
\toprule
& \textbf{IHdev-Acc.} & \textbf{IHtest-Acc.} \\
\midrule
RoBERTa-large & 73.07 ($\pm$.45) & 68.69 ($\pm$.56) \\ 
\midrule
+ RGCN     & 72.69 ($\pm$.19) & 68.41 ($\pm$.66) \\
+ GconAttn & 72.61 ($\pm$.39) & 68.59 ($\pm$.96) \\
+ KagNet   & 73.47 ($\pm$.22) & 69.01 ($\pm$.76) \\
+ RN       & 74.57 ($\pm$.91) & 69.08 ($\pm$.21) \\
+ MHGRN    & 74.45 ($\pm$.10) & 71.11 ($\pm$.81) \\
\midrule
+ QA-GNN   & 76.54 ($\pm$.21) & 73.41 ($\pm$.92) \\
+ QA-GNN + MorsE & \textbf{77.67} ($\pm$.34) & \textbf{75.56} ($\pm$.21) \\
\bottomrule
\end{tabular}
\label{tab:qa}
\end{table}


\subsection{Model Analysis}
\label{sec:model-analysis}

\subsubsection{Ablation Study}

In this section, we do ablation studies on different components of \model.
To remove the meta-learning setting, we train \model~on the source KG directly but not on the tasks with support and query sets. 
For ablating the entity initializer, we initialize entity embeddings randomly for each task during meta-training and model adaption. 
For ablating the GNN modulator, we skip the procedure of the GNN modulator and use the initialized embeddings from the entity initializer as entity embeddings.
We conduct ablation studies on inductive link prediction tasks, and show the results of \model~with TransE, which is the most representative KGE method, in Table \ref{tab:ablation}. The results show that different components are important, and it's beneficial to model them jointly.
Moreover, we find that the meta-learning setting is essential for the performance, which indicates that simulating a set of tasks for meta-training benefits the model generalization a lot.
Furthermore, the result significantly decreases when removing the GNN modulator on WN18RR (v1), and it's reasonable since the number of relations in WN18RR is less than FB15k-237, so the entity embeddings only based on the entity initializer are naive.

\begin{table}[t]
\linespread{1.4}
\centering
\caption{MRR (\%) results of the ablation study.}
\begin{tabular}{lcc}
\toprule
& \textbf{WN18RR (v1)} & \textbf{FB15k-237 (v1)} \\
\midrule
MorsE & \textbf{66.01} & \textbf{61.93} \\
\midrule
w/o meta-learning & 54.29 & 56.13  \\
w/o entity initializer & 63.87 & 59.90  \\
w/o GNN modulator & 21.93 & 59.07  \\
\bottomrule
\end{tabular}
\label{tab:ablation}
\end{table}

\begin{figure}[t]
\centering
\includegraphics[scale=0.48]{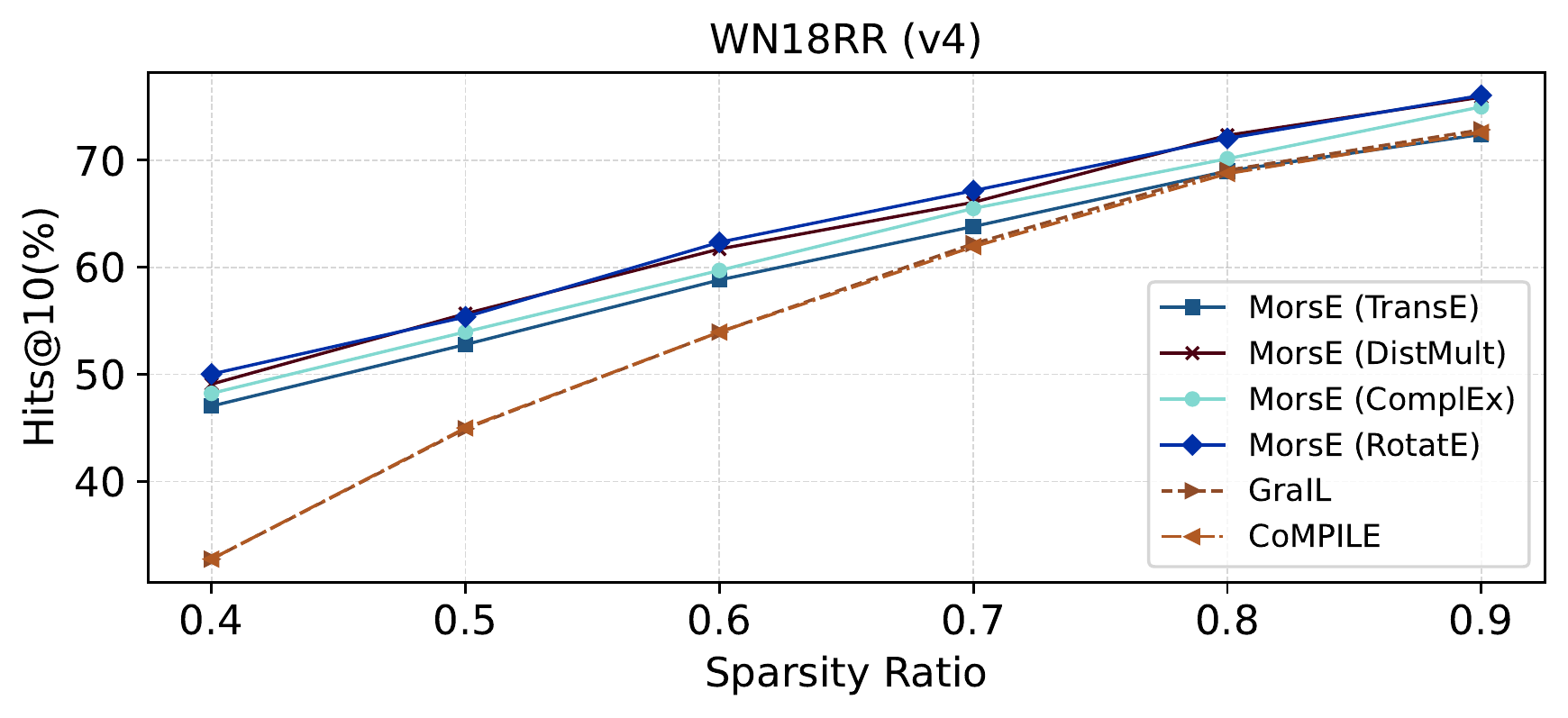}
\caption{Results on WN18RR (v4) with target KG remaining different ratios of triples (sparsity ratio).}
\label{fig:sparsity}
\end{figure}

\subsubsection{Target KG Sparsity Analysis}

The training regime of meta-learning is known to be beneficial to generalization on not only new tasks but also tasks with limited data \cite{maml}, so we believe that our proposed \model~can also obtain better performance on target KG with limited triples than other baselines for KGs in the inductive setting.
In the inductive link prediction task, we randomly delete triples in the target KG to make the target KG sparse and remain different ratios of triples. We call that ratio sparsity ratio.
We conduct inductive link prediction with various sparsity ratios on WN18RR (v4) since this dataset has enough triples in the target KG. 
From the results in Figure \ref{fig:sparsity}, we find that the performances for \model~with different KGE methods decrease less than GraIL and CoMPILE, as the target KG becomes sparser.
The results indicate that \model~is more robust to the sparsity of the target KG compared with baselines, which means that \model~can produce high-quality entity embeddings even though the number of triples in a KG is limited.

\section{Conclusion}

In this paper, we emphasize knowledge graph embedding as a general solution to represent knowledge graphs for in-KG and out-of-KG tasks and extend it to the inductive setting. Unlike current methods, which focus only on the completion task for KGs in the inductive setting, we raise a problem of inductive knowledge graph embedding, which can handle both in-KG and out-of-KG tasks for KGs in the inductive setting. 
To solve this problem, we propose a model \model~which considers transferring universal, entity-independent meta-knowledge by meta-learning.
Our experimental results show that the \model~outperforms existing state-of-the-art models on inductive link prediction tasks. Moreover, \model~also achieves the best results on the question answering task with KGs in the inductive setting.
Further model analyses indicate that components in \model~are essential, and \model~is more robust to the sparsity of the target KG than other baselines. 
In the future, we plan to explore more inductive settings for tasks related to knowledge graphs and evaluate our proposed model on more applications. 
\begin{acks}
This work is funded by NSFC U19B2027/91846204, Zhejiang Provincial Natural Science Foundation of China (No. LGG22F030011), Ningbo Natural Science Foundation (2021J190), and Yongjiang Talent Introduction Programme (2021A-156-G).
\end{acks}

\bibliographystyle{ACM-Reference-Format}
\bibliography{sample-base}


\appendix

\section{Dataset Statistics}
\label{app:table-statistic-lp}

We show dataset statistics for the inductive link prediction task in Table \ref{tab:statistic-lp}, and due to the space limitation, we abbreviate the name for each dataset; for example, W1 denotes \underline{W}N18RR (v\underline{1}).

\input{table-statistic-lp}

\end{document}

%% file: table-results-lp.tex
\begin{table*}[t]
\centering
\caption{Hits@10 (\%) of link prediction for KGs in the inductive setting. Results of baselines are taken from \citet{compile}. Bold numbers denote the best results of all models and underline numbers denote the best results of baselines.}
\begin{tabular}{lllll|llll|llll}
\toprule
& \multicolumn{4}{c}{\textbf{WN18RR}} & \multicolumn{4}{c}{\textbf{FB15k-237}} & \multicolumn{4}{c}{\textbf{NELL-995}} \\ 
\cmidrule(lr){2-5} \cmidrule(lr){6-9} \cmidrule(lr){10-13} 
& \multicolumn{1}{c}{\textbf{v1}} & \multicolumn{1}{c}{\textbf{v2}} & \multicolumn{1}{c}{\textbf{v3}} & \multicolumn{1}{c}{\textbf{v4}} & \multicolumn{1}{c}{\textbf{v1}} & \multicolumn{1}{c}{\textbf{v2}} & \multicolumn{1}{c}{\textbf{v3}} & \multicolumn{1}{c}{\textbf{v4}} & \multicolumn{1}{c}{\textbf{v1}} & \multicolumn{1}{c}{\textbf{v2}} & \multicolumn{1}{c}{\textbf{v3}} & \multicolumn{1}{c}{\textbf{v4}} \\ \midrule
Neural-LP 
& 74.37 & 68.93 & 46.18 & 67.13 
& 52.92 & 58.94 & 52.90 & 55.88 
& 40.78 & 78.73 & 82.71 & \underline{80.58} \\
DRUM 
& 74.37 & 68.93 & 46.18 & 67.13 
& 52.92 & 58.73 & 52.90 & 55.88 
& 19.42 & 78.55 & 82.71 & \underline{80.58} \\
RuleN 
& 80.85 & 78.23 & 53.39 & 71.59 
& 49.76 & 77.82 & \underline{87.69} & 85.60 
& 53.50 & 81.75 & 77.26 & 61.35 \\ \midrule
GraIL 
& 82.45 & 78.68 & 58.43 & 73.41 
& 64.15 & 81.80 & 82.83 & \underline{89.29} 
& \underline{59.50} & 93.25 & 91.41 & 73.19 \\
CoMPILE 
& \underline{83.60} & \underline{79.82} & \underline{60.69} & \underline{75.49} 
& \underline{67.64} & \underline{82.98} & 84.67 & 87.44 
& 58.38 & \textbf{\underline{93.87}} & \textbf{\underline{92.77}} & 75.19 \\ 

\midrule

MorsE (TransE)
& 81.45 & 78.84 & 67.30 & 76.46
& \textbf{84.74} & \textbf{96.31} & \textbf{95.79} & \textbf{96.43}
& 51.76 & 83.81 & 89.28 & \textbf{82.42} \\ 

MorsE (DistMult)
& 83.74 & 79.98 & 66.06 & 79.33 
& 82.51 & 94.93 & 94.93 & 95.75  		
& \textbf{66.56} & 79.40 & 84.35 & 54.06  \\  

MorsE (ComplEx)
& 83.54 & 81.39 & 67.28 & 79.43 		
& 79.95 & 92.78 & 93.64 & 95.07 		
& 64.98 & 76.92 & 82.95 & 42.58   \\ 

MorsE (RotatE)
& \textbf{84.14} & \textbf{81.50} & \textbf{70.92} & \textbf{79.61}
& 83.17 & 95.67 & 95.69 & 95.89
& 65.20 & 80.70 & 87.67 & 53.44 \\

\bottomrule
\end{tabular}
\label{tab:link-prediction-rst}
\end{table*}

%% file: table-statistic-lp.tex
\begin{table}[th]
\renewcommand\arraystretch{1.1}
\centering
\caption{Statistics of various versions of WN18RR (W), FB15k-237 (F) and NELL-995 (N).
We show the number of relations (\#rel), entities (\#ent) and triples (\#tri) in the source KG, and the number of relations, entities, triples as well as test triples (\#test) in the target KG.}
\begin{tabular}{lrrr|rrrr}
\toprule
& \multicolumn{3}{c}{\textbf{Source KG}} & \multicolumn{4}{c}{\textbf{Target KG}} \\
\cmidrule(lr){2-4} \cmidrule(lr){5-8}
& \textbf{\#rel} & \textbf{\#ent} & \multicolumn{1}{r}{\textbf{\#tri}} & \textbf{\#rel} & \textbf{\#ent} & \textbf{\#tri} & \textbf{\#test} \\
\midrule
W1 & 9 & 2,746 & 6,678 & 8 & 922 & 1,618 & 188 \\
W2 & 10 & 6,954 & 18,968 & 10 & 2,757 & 4,011 & 441 \\
W3 & 11 & 12,078 & 32,150 & 11 & 5,084 & 6,327 & 605 \\
W4 & 9 & 3,861 & 9,842 & 9 & 7,084 & 12,334 & 1,429 \\
\midrule
F1 & 180 & 1,594 & 5,226 & 142 & 1,093 & 1,993 & 205 \\
F2 & 200 & 2,608 & 12,085 & 172 & 1,660 & 4,145 & 478 \\
F3 & 215 & 3,668 & 22,394 & 183 & 2,501 & 7,406 & 865 \\
F4 & 219 & 4,707 & 33,916 & 200 & 3,051 & 11,714 & 1,424 \\
\midrule
N1 & 14 & 3,103 & 5,540 & 14 & 225 & 833 & 100 \\
N2 & 88 & 2,564 & 10,109 & 79 & 2,086 & 4,586 & 476 \\
N3 & 142 & 4,647 & 20,117 & 122 & 3,566 & 8,048 & 809 \\
N4 & 76 & 2,092 & 9,289 & 61 & 2,795 & 7,073 & 731 \\
\bottomrule
\end{tabular}
\label{tab:statistic-lp}
\end{table}